\documentclass[letterpaper, 10 pt, conference]{ieeeconf}  

\IEEEoverridecommandlockouts                              

\usepackage{graphicx} 
\usepackage{float}
\usepackage{mathptmx} 
\usepackage{amsmath} 
\usepackage{amssymb}  

\title{\LARGE \bf
Multi-Task Learning of Active Fault-Tolerant Controller for Leg Failures in Quadruped robots
}

\author{Taixian Hou$^{1}$, Jiaxin Tu$^{1}$, Xiaofei Gao$^{2}$, Zhiyan Dong$^{3}$, Peng Zhai$^{1,*}$, Lihua Zhang$^{4,*}$
\thanks{$^{1}$Taixian Hou, Peng Zhai, Jiaxin Tu, are with the Academy for Engineering and Technology, Fudan University, Shanghai 200433, China
        {\tt\small txhou21@m.fudan.edu.cn; pzhai@fudan.edu.cn; jxtu22@m.fudan.edu.cn}}%
\thanks{$^{2}$Xiaofei Gao is with Beijing Jingcheng Zhitong Robotics Technology Co., Beijing, China
        {\tt\small gaoxiaofei@inter-smart.com}}%
\thanks{$^{3}$Zhiyan Dong is with the Ji Hua Laboratory, Foshan 200433, China
        {\tt\small dongzhiyan@fudan.edu.cn}}%
\thanks{$^{4}$Lihua Zhang is with the Engineering Research Center of AI and Robotics, Shanghai, China
        {\tt\small lihuazhang@fudan.edu.cn}}%
\thanks{*Corresponding Author.}
}

\begin{document}

\maketitle
\thispagestyle{empty}
\pagestyle{empty}

\begin{abstract}

Electric quadruped robots used in outdoor exploration are susceptible to leg-related electrical or mechanical failures. Unexpected joint power loss and joint locking can immediately pose a falling threat. Typically, controllers lack the capability to actively sense the condition of their own joints and take proactive actions. Maintaining the original motion patterns could lead to disastrous consequences, as the controller may produce irrational output within a short period of time, further creating the risk of serious physical injuries. This paper presents a hierarchical fault-tolerant control scheme employing a multi-task training architecture capable of actively perceiving and overcoming two types of leg joint faults. The architecture simultaneously trains three joint task policies for health, power loss, and locking scenarios in parallel, introducing a symmetric reflection initialization technique to ensure rapid and stable gait skill transformations. Experiments demonstrate that the control scheme is robust in unexpected scenarios where a single leg experiences concurrent joint faults in two joints. Furthermore, the policy retains the robot's planar mobility, enabling rough velocity tracking. Finally, zero-shot Sim2Real transfer is achieved on the real-world SOLO8 robot, countering both electrical and mechanical failures.

\end{abstract}

\begin{figure*}[t]
    \centering
    \includegraphics[width=\linewidth]{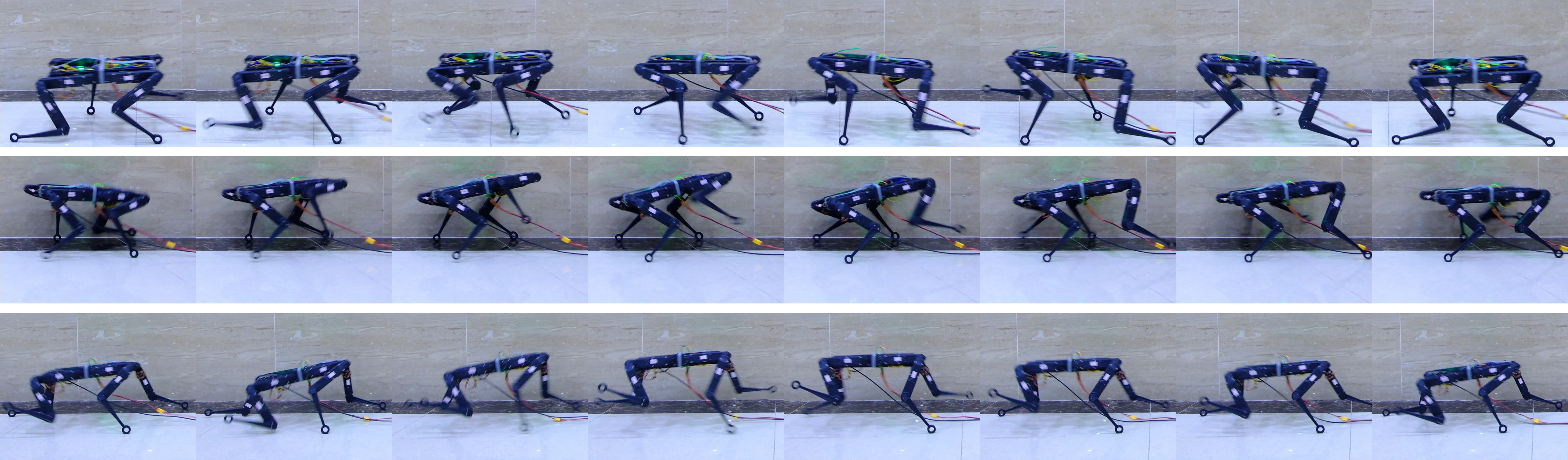}
        
        
    \caption{Sim2Real tasks were conducted on SOLO8 robots under different leg statuses: (Top) Health, (Mid) Limit, and (Low) Weak. 'Limit' and 'Weak' denote scenarios where locking(we simulate with larger PD) and electrical faults occur in two joints within left-front leg, respectively. Our control scheme demonstrates diverse gait skills, adapting to leg joint faults while accurately tracking human commands.}
    \label{fig:1}
\end{figure*}
\section{INTRODUCTION}

In pursuit of unlocking the full potential of quadruped robot locomotion and enhancing their adaptability in complex real-world environments, researchers have persistently invested their efforts in the domains of robust control and intricate motion generation \cite{RN26} . However, early quadruped robots were severely limited in their capacity to produce complex movements due to constraints stemming from both mechanical properties and control methodologies \cite{RN21, RN30}.

In recent years, besides model-based solutions \cite{RN18, RN31}, the intersection of quadruped robotics and reinforcement learning has witnessed a surge in modern research endeavors, aimed at augmenting traversal ability across rugged terrains \cite{RN11,RN32} or challenging stairs \cite{RN28}, imitating natural behaviors \cite{RN29, RN17, RN13}, expanding the range of motion skills \cite{RN42}, improving adaptability \cite{RN12, RN51}, and fortifying resistance to disturbances while accelerating training speed \cite{RN23, RN15}. Contemporary quadruped algorithms have demonstrated remarkable capabilities in navigating natural landscapes and performing tasks in unstructured outdoor environment, even in scenarios with limited or unreliable sensory information, such as wilderness mountain climbing \cite{RN16}. While advanced control capabilities are undeniably critical for wilderness exploration, ensuring the safety and stability of these robots in malfunctions is equally indispensable. Preemptively equipping robots with effective policies to autonomously handle unforeseen failures can significantly reduce labor-intensive maintenance costs and mitigate both direct and indirect economic losses.

Drawing inspiration from the remarkable adaptability and control abilities inherent in the natural world's animals, it becomes evident that many terrestrial quadrupeds, facing leg injuries or leg loss, possess the capacity to adjust their strategies and adapt to their altered physical configurations.

In this paper, we aspire to empower quadruped robots with the ability to actively perceive and adapt to both electrical and mechanical failures, thereby enhancing their resilience in challenging outdoor environments. 

This work contributed on:  (1) We propose a layered fault-tolerant control scheme suitable for quadruped robots to actively perceive joint failures, addressing the common leg-related issues encountered during outdoor exploration. This work extends prior fault-tolerant studies on quadruped robots by covering electrical failure scenarios that were rarely explored. (2) The training is 
established on an online multi-task learning architecture, leveraging task incentives for robots to master various gait skills. Additionally, we introduced a reflection initialization technique to simulate different initial states of fault scenarios, thereby enhancing the capability to transition from different leg statuses. (3) Sim-to-real evaluation of three leg statuses was conducted on the physical SOLO8 robot. By adjusting gait skills, the robot is capable of maintaining mobility and respondding to human commands.

\section{RELATED WORK}

\subsection{Fault-Tolerant Control}

Our aspiration is for quadruped robots to venture beyond the laboratory and be utilized in outdoor environments. Filling the blank of self-protection mechanisms and post-failure skills for quadruped robots. Some fault-tolerant control strategies for hexapod robots, including joint locking scenarios, were built upon model-based approaches \cite{RN33,RN34}. Nevertheless, these approaches tend to struggle with adapting to the influences of unknown environments, balancing gait becomes more challenging once leg-related joint issues are introduced.

\subsubsection{\textbf{Joint Locking Challenge}} In recent years, research on model-based fault-tolerant control for quadruped robots has focused on addressing joint locking issues. There are methods based on G(F) \cite{RN39} and gait design using Whole-Body Control (WBC) \cite{RN7}. These methods are tailored to specific robots, lacking generalization. Moreover, as designing gaits for failures and modeling different scenarios requires substantial manual analysis, such algorithms have not seen widespread real-world deployment, limiting their practical applicability.

Reinforcement Learning approach: Early reinforcement learning fault-tolerant policies \cite{RN40} struggled with sim-to-real discrepancies and failed in real-world experiment. A recent approach by Liu et al. \cite{RN6} tackled the joint locking problem using a teacher-student network and a well-defined environment with 3D-printed parts simulating joint locking. The work achieved Sim2Real testing on a real A1 quadruped robot which maintained forward locomotion and heading direction. However, this work solely focused on single joint locking scenario and constrained the post-failure robot's movement to a fixed speed, losing the ability for planar locomotion and tracking human velocity commands.

\subsubsection{\textbf{Joint Power Loss Challenge}} Few studies research on joint power loss scenarios. Recent work explores three-legged dynamic walking of quadruped robots through model-based approaches \cite{RN41}. In contrast, our controller aims to robust leg electrical fault tolerance, naturally developing jumping capabilities. Additionally, our strategy addresses the stable transition of gait skills at the moment of abrupt electrical failure, regardless of the robot's current joint states.

In this paper, we extend our focus from single joint failures to the scenario of dual joint failures within a single leg. Addressing the potential simultaneous failure of both the hip and knee joints significantly complicates fault-tolerant control. We also combined the challenges of both two types of leg failure and conducted a complete study on gait skill transformation. Finally, our strategy retains planar mobility and the capabilities of adhering to human velocity commands.

\subsection{Hierarchical settings and Multi-Task Reinforcement Learning}

In practice, we observed that the Solo8 robot, with only 8 degrees of freedom, lacks the flexibility compared to mainstream 12-DOF quadruped robots like A1, Go1, and Anymal. Under similar reward settings, it struggled to acquire normal gaits in different directions. Despite our efforts, after referencing the work of previous experiment on SOLO8 \cite{RN17} and SOLO12 \cite{RN43} and manually adjusting training parameters, we found that the gaits for tracking forward and backward commands remained distinct. This insight led us to the realization that a single model struggles to generalize across different behavior distributions.

Practical experience tells us that the output by the fault-tolerant policy will further deviate from the original distribution. A Single model is hard to master distant distributions, as we will elaborate in the results discussion.

One approach is to draw inspiration from fields like NLP and increase the model's capacity to enhance its ability to differentiate special skills and achieve diversified outputs \cite{RN47,RN48}. In the pre-training process of motion capture imitation learning, a larger model was used, and a lower-level controller with mutual information objectives was trained. Subsequently, the parameters of the lower-level controller were frozen, and the upper-level controller was trained using reinforcement learning to address multiple tasks.

Another perspective stems from multi-task learning, where we aim to train a joint hierarchical strategy. Several layers within it can operate without parameter sharing, enabling the model to maintain a smaller footprint. Typically, this approach is predominantly employed in the pre-training process of offline reinforcement learning \cite{RN22}. We posit that this limitation primarily arises from the historical emphasis of prior online learning techniques on single or limited environments within a single thread.

By harnessing the parallel training capabilities in IsaacGym \cite{RN44} across thousand of environments, and appropriately grouping environments with different tasks, we concurrently train on various tasks and swiftly gather rich samples. Our approach offers two advantages: firstly, it allows us to maintain a small and refined model, coupled with an additional fault discriminator, achieving distinct behaviors. Secondly, leveraging multiple environments partitioning, robots rapidly acquire various skills without relying on external prior knowledge or distinct reward function settings. Compared to large models and motion capture imitation, our approach is faster and more user-friendly.

\section{METHOD}

\subsection{Overview} 

Since the controller must be able to adapt to different leg status and present dynamic changes in gait patterns, developing a single strategy that excels across all tasks becomes challenging. We employ a hierarchical multi-task scheme with joint task policy $\pi(a|s):=\pi_{\mu}(z|s)\pi_i(a_i|z)$ to control the robot, where $\pi_{\mu}(z|s)$ represents a front policy(network) with common parameters, and $\pi_i(a_i|z)$ denotes the task specified low-level policy. The set of task-specific policies is denoted as $\{\pi_1(a_1|z),\pi_2(a_2|z),\pi_3(a_3|z),...,\pi_N(a_N|z)\}$, and these policies do not share parameters. In this work, we have trained three task-policies correspond to various leg statuses: health, limit, and weak. These correspond to environments without joint faults, with joint locking, and with power loss scenarios, respectively. 

\begin{figure}[t]
  \centering
  \includegraphics[width=0.47\textwidth]{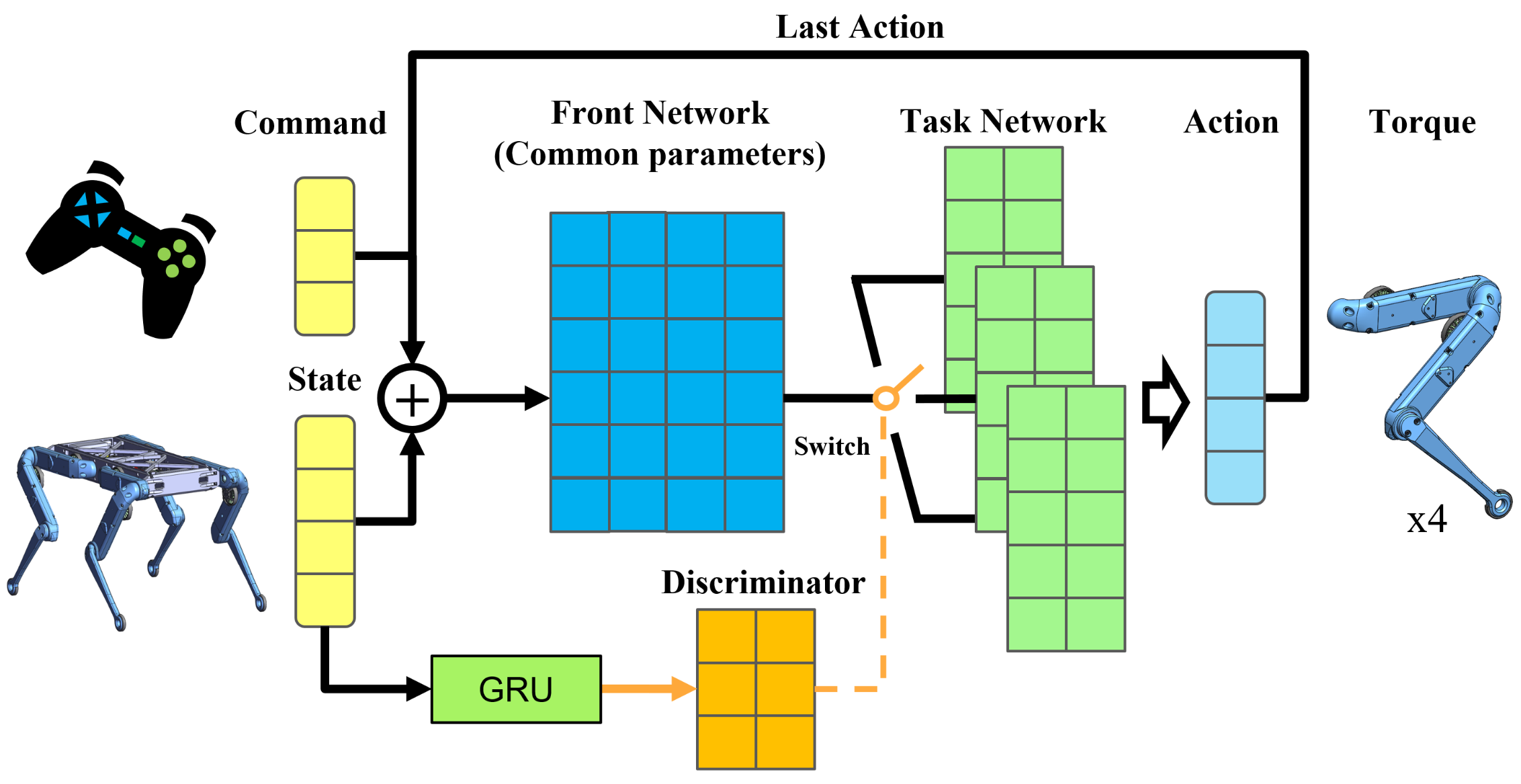}
  \caption{Scheme of the different controller components}
  \label{fig:2}
\end{figure}

Our control scheme is shown in Fig. \ref{fig:2}. Observation inputs encompass human commands, robot proprioceptive information (including IMU), and the previous actions. Joint states of the robot, encoded by GRU units, are fed into a fault discriminator, which aids the robot in switching low-level task policies in different scenarios.

\subsection{Training architecture}

We define this training as a multi-task Markov Decision Process (MDP) $M=(S,A,P,r,\{R_i,i\}{i=1}^N)$. 
Here, $S$ represents the state space, $A$ denotes the action space, and  $P(s' | s, a)$ is the dynamic transition function. 
The index $i$ represents the task number, with a total of $N$ tasks. While the reward functions for different environments remain consistent across all tasks, the actual reward, denoted as $R_i$, may still be influenced by the specific setup of the environment. Utilizing the Proximal Policy Optimization (PPO) algorithm, we aim to find the optimal policy $\pi^*(a|s,\cdot):=\arg\max\pi \mathbb E_{i\sim [N]}\mathbb E_{\tau\sim \pi(i)}\sum\gamma R(\tau(i))$, where $\gamma$ denotes the discount factor, and $\tau(i)$ represents the trajectory of task $i$. Similar to the policy referred in last section, the critic function $V(s, i)$ possesses a similar hierarchical structure $V(s,i):=V_{\mu}(z|s,i)V_i(z)$, used to estimate the value function of the policy $\pi(\cdot|\cdot,i)$. The task-specific low-level critic $V_i(z)$ is solely employed for evaluating task environment $i$ and is updated by data sampled from corresponding joint task policy $\pi(a|s,i)$. The training is divided into two stages:

\subsubsection{\textbf{First training stage}}

We propose an architecture for multi-task reinforcement learning in parallel environments. To enable the robot to learn more general gaits, in the first stage, we partition the environments in a ratio of 2:1:1. Through IsaacGym simulator, we simultaneously train 2048 normal environments (health), 1024 electrical fault environments (weak) and 1024 mechanical fault environments (limit). Leveraging the substantial advantage of parallel training in terms of speed, we achieve online multi-task learning. The action in environment $i$ is generated by the joint task policy $\pi(a|s,i)$. Data collected from environments of the same task is exclusively used for updating the corresponding low-level policy and critic, to acquire different gait skills.

\begin{figure}[h]
  \centering
  \includegraphics[width=0.47\textwidth]{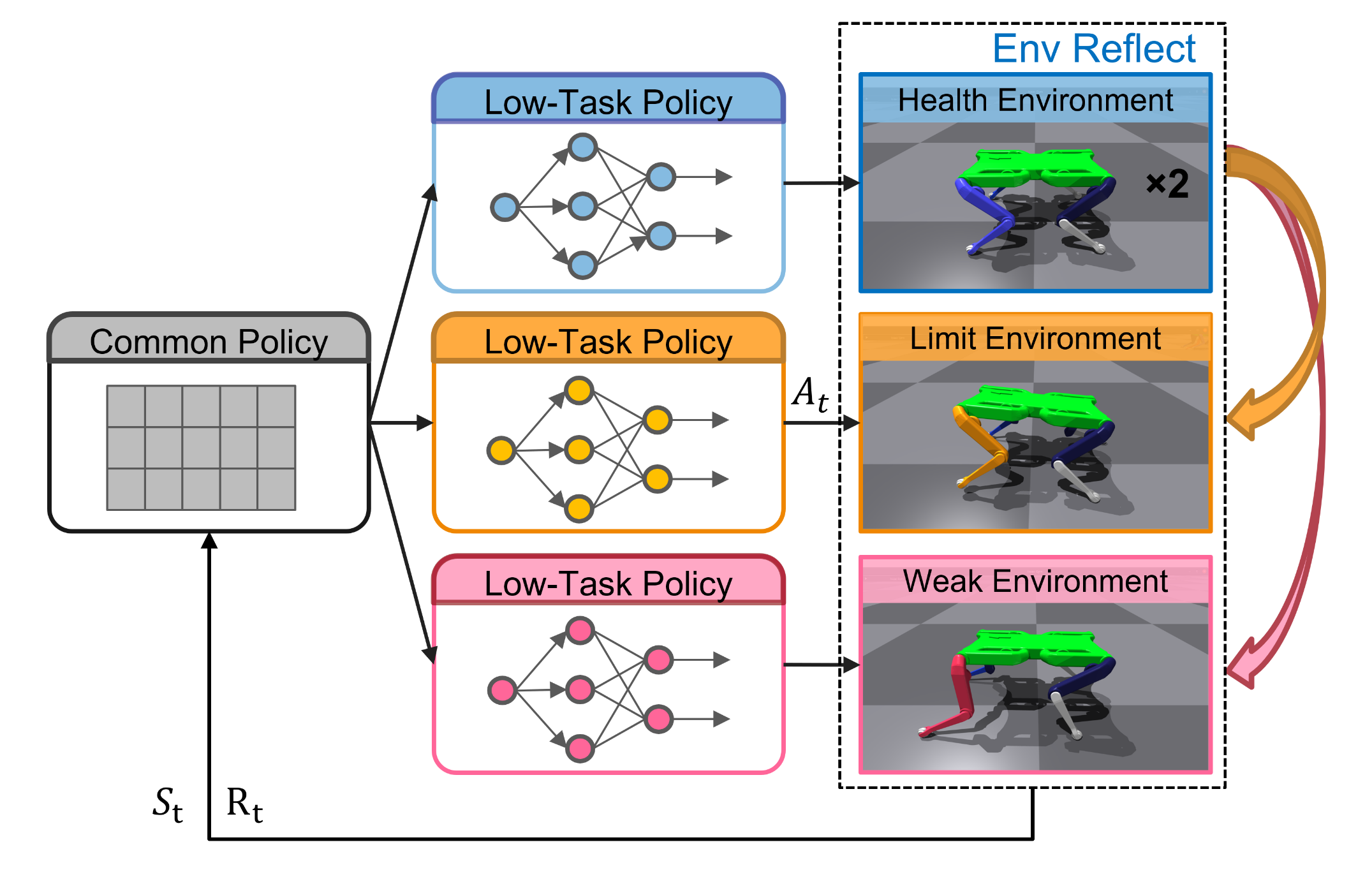}
  \caption{Architectural during the first training stage. ('Limit' denotes mechanical locking, 'Weak' denotes electrical power loss). Symmetric 'Health' states were reflected into 'Limit' and 'Weak' environments. This technique fosters the gaits transfer performance.}
  \label{fig:3}
\end{figure}

To facilitate the trained policy for seamless transitions from a random health leg joint state to a specific fault scenario, we introduce a symmetric environment reflection initialization method. As depicted in Fig. \ref{fig:3}. Each first step of a faulty environment's episode, reflected states from symmetric health environment. In contrast to approach described in \cite{RN6}, where a random time in the latter half of an episode is chosen for the quadruped robot to experience locking fault. While the setup involves several steps of leg state transition, it can lead to a more cautious progression of training, even lead to odd action policy when training a single policy within two leg statuses in episode. Conversely, in the approach we propose, the faulty environment group only copy the states at the beginning of an episode. This approach fosters the ability for smooth transitions and preserves the potential for the robot to acquire diverse gait skills.

\subsubsection{\textbf{Second training stage}}

To further train the robot's state transition capability, in the second stage, we introduced random fault times $T^{f}_{1}\sim T^{f}_{n}$ for each agent, where $T^f_i$ can occur at any time within an episode. Before the first $T^f_{1}$, the robot operates in a healthy environment, and then randomly enters either joint power-loss or joint lock. Subsequent $T^f_{i}$ can select randomly, including health status, implying that faults may recover after a brief occurrence. This allows us to simulate more potential real-world scenarios.

Simultaneously, to enable the robot to actively detect faults and automatically switch to fault-tolerant gaits, we also trained a fault discriminator capable of perceiving leg joint faults. The discriminator is composed of a Multi-Layer Perceptron (MLP), taking encoded short-term states $y_t$ as input, outputs $\mathbf{u} \in \mathbb{R}^3$, denotes to the probabilities of health, limit, and weak status.

The fault discriminator, front network, and all low-level policies are queried at 50Hz. Compared to physical machines in the real-world, frequency of joint policies execution remains consistent.

\begin{figure}[!h]
  \centering
  \includegraphics[width=0.45\textwidth]{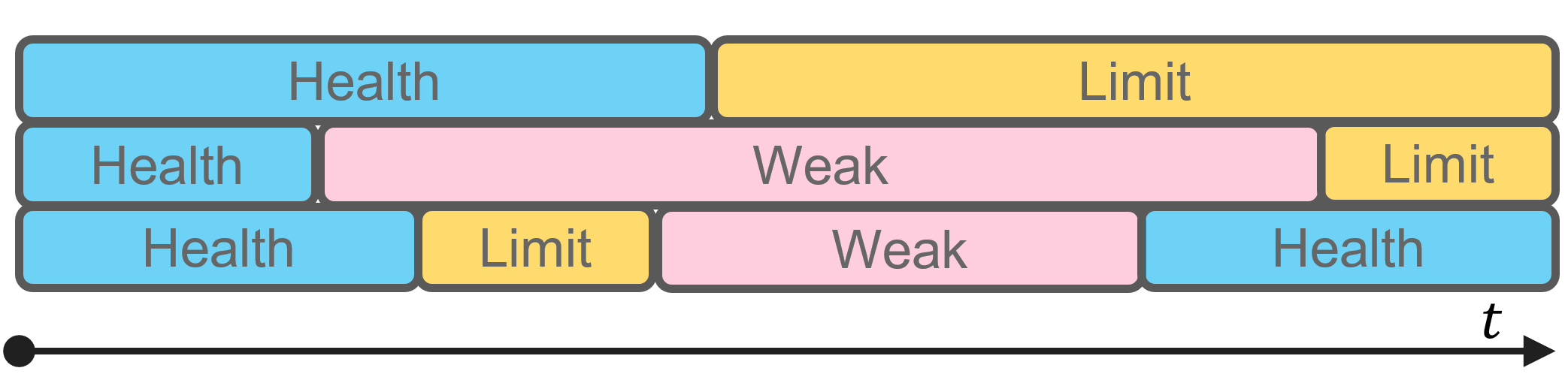}
  \caption{Leg status randomly transforms during the second training stage. Three example episodes are given here.}
\end{figure}

\subsection{Environment and Learning Configuration}

\textbf{Observations and States}:
Our observation space is composed of spatial attitude quaternion, the position and velocity states information of the robot's eight joints, the action output from the previous round of policy, and the forward and yaw velocity commands from the remote controller, forming a 30-dimensional observation space.

Attitude quaternion from IMU offer similar information to gravity vectors, but they also provide absolute yaw orientation. In line with prior research, we have the flexibility to select one that aligns best with our task.

\textbf{Action}:
The action output provides target angular deviations for the eight motors from their initial angular positions. It is generated by the joint task policy $\pi(a|s,i):=\pi_{\mu}(z|s)\pi_i(a_i|z)$. The final target positions, after simulation, are used to compute the torque output at 200Hz via $\tau = K_P(q_0 + a -q)-K_d(\dot q)$. The frequency, while not as high as in practical applications, is sufficient for simulating the crucial movements of the robot. The SOLO8 robot which we test in the real world  employs a higher PD control frequency at 10000HZ.

\textbf{Reward Functions}:
Following the intricate reward style outlined in \cite{RN15,RN43}, we have incorporated special foot-clearance and foot-slip rewards for SOLO8 robots. Given the mechanical structure of SOLO8 which lacks hip abduction/adduction (HAA) joints, it is unable to walk in lateral directions. The foot-clearance reward related to foot velocity focuses on robot's foot height while stepping forward. Without this term, robots might exhibit odd actions with increased friction on the floor.

Apart from foot height considerations, we have introduced a straightforward penalty for slipping in the reward functions. Here, we abbreviate other reward functions and emphasize the critical terms for SOLO8 robots. These reward setting are defined as:

\begin{equation}
r_{clearance} = c_{clearance}\sum_{j=1}^4 |p_{z,j}-p_{z}^{m}|\sqrt{|p_x|}
\end{equation}

where $j$ denotes the foot number, $p_z^{m}$ is the expected foot height, $p_{z,j}$ is the foot center height, $p_x$ is the forward foot speed, and $c_{clearance}$ is the clearance penalty scale.

\begin{equation}
    r_{slip} = c_{slip}\sum_{j=1}^4 I_j {\parallel p_{xy} \parallel}^2
\end{equation}

where $I_j$ is a filtered indicator of whether foot $j$ is in contact with the ground, $p_{xy}$ is the foot's plane speed, and $c_{slip}$ is the slip penalty weight.

To encourage early exploration, we implemented a reward curriculum approach where the weight of the penalty term $R_{pen}$ progressively increases over the course of training. Here, $i$ represents the current iteration of training, and $N$ denotes the total number of iterations. Additionally, the total reward is clipped to a positive value, ensuring that individual penalties take effect only when the reward $R_{sum}$ is positive.

\begin{equation}
R_{sum}(i) = \max(\frac{i}{N}R_{pen} + R_{rew} , 0)
\end{equation}

\textbf{Domain Randomization}:
Domain randomization parameters are used to enhance the robustness of simulation-to-real transfer. This primarily focuses on the errors in simulating different frictions and the overall robot mass in reality. It also provides a range of speeds for simulating forward and yaw commands from the remote control. Additionally, due to the necessity of manually adjusting and calibrating the initial position of the SOLO8 robot, a random initial state can also be utilized.

\begin{table}[!ht]
  \centering
  \caption{Domain Randomization Parameters}
  \begin{tabular}{|c|c|c|}
    \hline
    Randomization            & $Lower\ bound$ & $Upper\ bound$ \\
    \hline
    Friction                 & 0.6   & 1.5    \\
    Mass                     & -0.2  & 1      \\
    Linear velocity commands & -0.   & 1      \\
    Angular velocity commands& -0.5  & 0.5    \\
    Initial state            & -0.1  & 0.1    \\
    \hline
  \end{tabular}
\end{table}

\section{RESULTS}

The aim of following experiments is to investigate:

\begin{enumerate}
  \item Whether our training process enables agents to learn diverse gait skills that deviate from the original distribution, and if there is enough "distance" between three gaits states.
  \item In comparison to agent only trained in base environment and agent trained in all the environment (health, limit, weak) but without our architecture, whether our approach can increase the survival rates when robots encounter joint electrical or mechanical failures. Does our reflection technique increase performance?
  \item How much of the planar locomotion ability is retained? Can the robot be tracking along the forward velocity and direction?
  \item How do the generated gaits of our method look? Can our method effectively transfer to sim2real tasks, and exhibit diverse gait skills on a real quadruped robot?
\end{enumerate}

\subsection{Experiment on Gait skills distribution}

In the distribution experiments, we collected data on the states and actions of 120 steps generated by the joint policies in different task-type environments. There are three sets of environments, each corresponding to a leg status (health, weak, limit), with six agents in each set. From an intuitive perspective, the strategy naturally formed three different gaits. The quadruped robot learned clear and natural footsteps in the health status, three-legged jumping with two joints electrical fault, and wiggly walking by adjusting the body equilibrium with two joints locking. Watch our videos for an intuitive view.

\begin{figure}[t]
    \centering
    \begin{minipage}{0.15\textwidth}
        \centering
        \includegraphics[width=\linewidth]{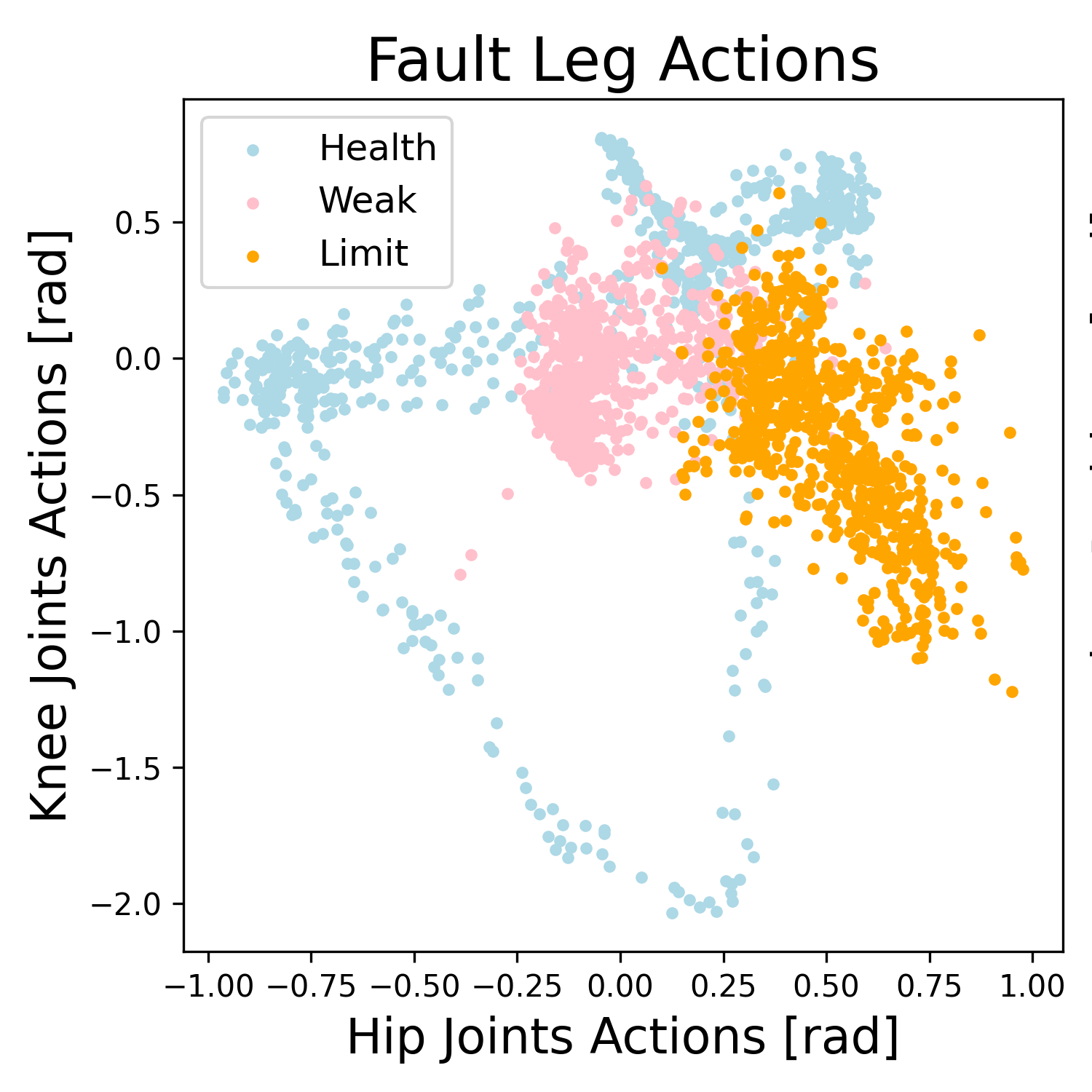}
        (a)
    \end{minipage}
    \hfill
    \begin{minipage}{0.15\textwidth}
        \centering
        \includegraphics[width=\linewidth]{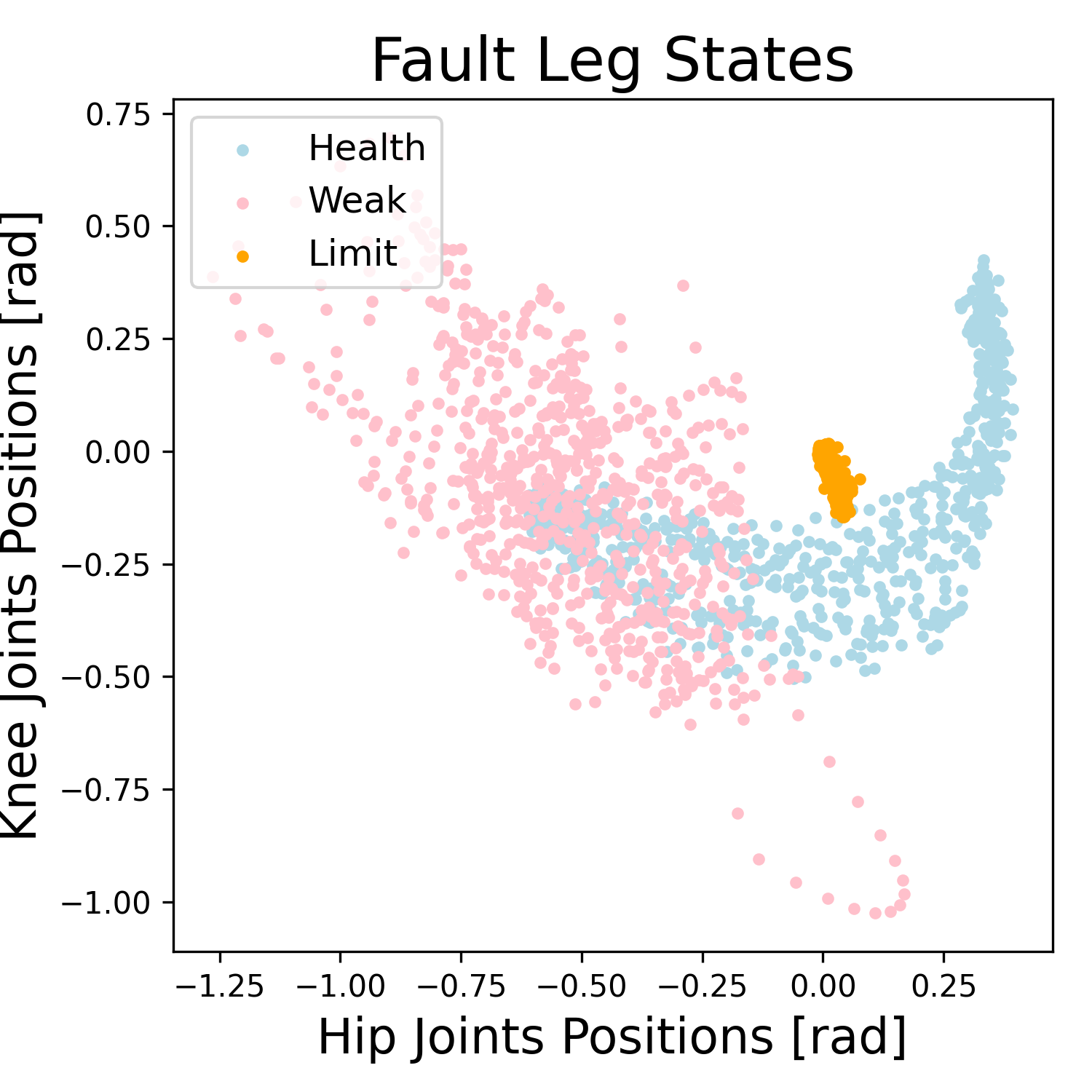}
        (b)
    \end{minipage}
    \hfill
    \begin{minipage}{0.15\textwidth}
        \centering
        \includegraphics[width=\linewidth]{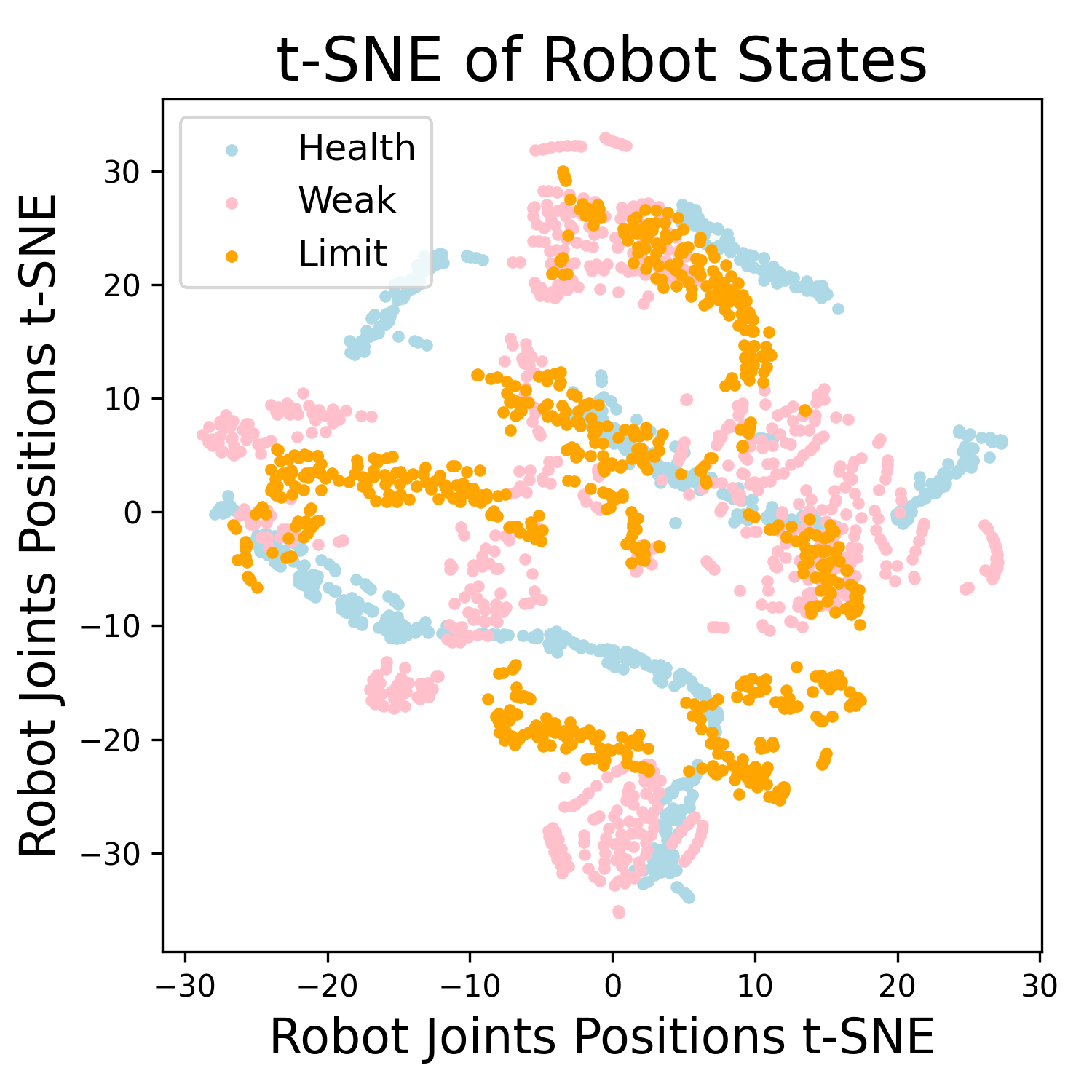}
        (c)
    \end{minipage}
    \caption{Joints data from 120-steps trajectories of 6 quadruped agents. (a) and (b) show the actions and states of Hip Joint and Knee Joint within fault-leg. (c) t-SNE visualization of the robot(all 8 joints) state space.}
    \label{fig:5}
\end{figure}

As shown in Fig. \ref{fig:5}(a) and (b) show the actions and joint states of the single-leg Hip Joint and Knee Joint of the robot, where actions in the health environment exhibit periodicity within a certain range. In contrast, actions in the weak and limit environments are more concentrated. Fig. \ref{fig:5}(c) t-SNE visualization of the robot's state space reveals distinct distribution of whole joints.

\subsection{Comparison of Survival Rates}

In this set of experiments, we aim to figure out the performance of our approach.We defined a robot falling within an episode as "death," while if the episode continued until the final step, it was considered "survival." We designed a series of experiments to evaluate the robot's survival rate. 

The experimental environments were divided into 16 groups, each consisting of 100 environments, totaling 1600 environments. In each environment, the quadruped robot received a forward velocity command of $V_x \sim (0.4\pm 0.1)$. The fault settings in three separate experiments were similar to our second training stage. In the healthy environment experiment, the robot encountered no troubles throughout the entire 8-second episode. In the weak and limit experiment, the quadruped robot experienced leg faults after a random fault time $T_f$. 

To compared with, we trained a normal(baseline) policy and mix policy \cite{RN6}. These strategies lacked a hierarchical structure, and they were trained exclusively in either healthy or all leg status (health, limit, weak) environments. We compared them with our scheme in terms of survival rates. 

\subsubsection{\textbf{Analysis}}
\begin{figure}[!h]
  \centering
  \includegraphics[width=0.45\textwidth]{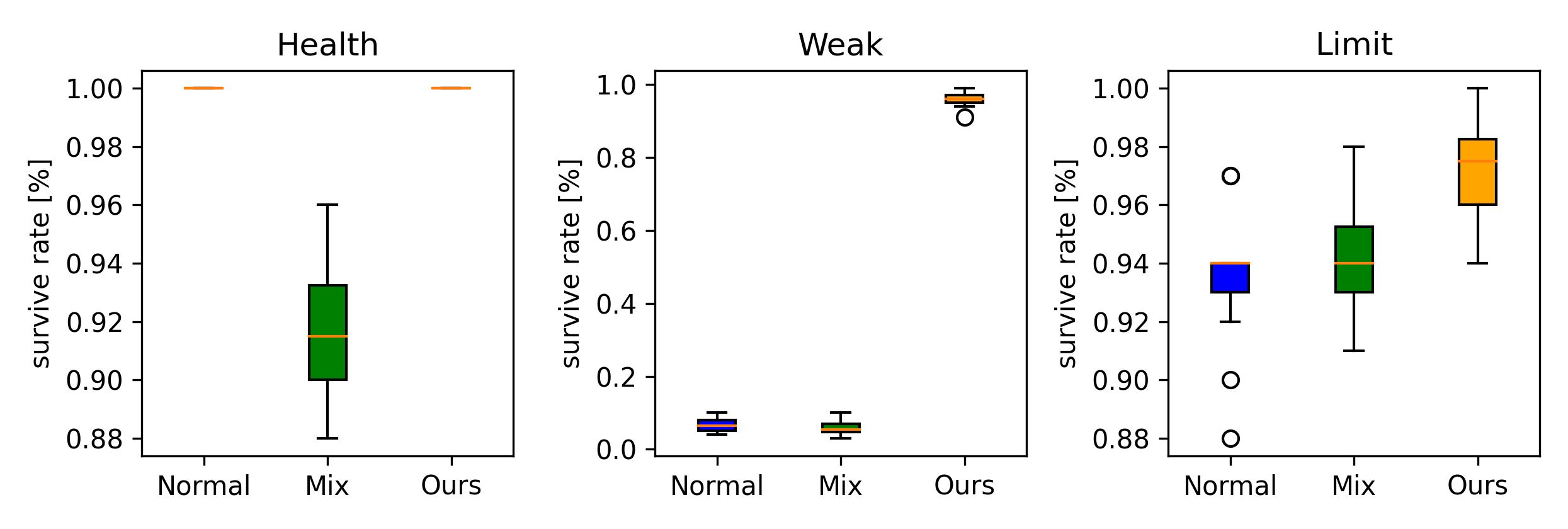}
  \caption{Survival rates of the robots in Health, Weak, Limit status environments. The normal and mix policies are compared with our approach.}
  \label{fig:6}
\end{figure}

The comparative experiment results are depicted in Fig. \ref{fig:6}. Our approach significantly improved survival rates under both joint faults, especially in the weak fault scenario. The agents learned effectively to handle combined electrical faults within two joints, utilizing the other three legs for jumping actions. The mixed strategy hardly learned anything in the weak environment. Although it slightly improved survival rates in the limit scenario, it compromised the survival rate in the health status. 

In our further investigation, we found that a single network struggled to master two significant skills within distant distributions. This resulted in 'Mix' policies using a minor scraping gait while dealing with both limit and health scenarios. While it showed slight improvement over the normal strategy in limit scenarios, the negative impact on natural gaits in the health status made it an unfavorable choice.

\subsubsection{\textbf{Ablation experiment}}
To verify the importance of the reflection technique, we conducted an additional set of ablation experiments within middle speed $V_x \sim (0.4\pm 0.1)$ (Weak, Limit). 
Moreover, to validate the necessity of the reflection technique and measure its improvement on gait transition capabilities, we introduced a high-speed velocity command test in the ablation experiment. The robots will face the scenario where joint failures occur during high-speed locomotion, posing a significant challenge to achieving stable gait transitions.

\begin{figure}[!h]
  \centering
  \includegraphics[width=0.45\textwidth]{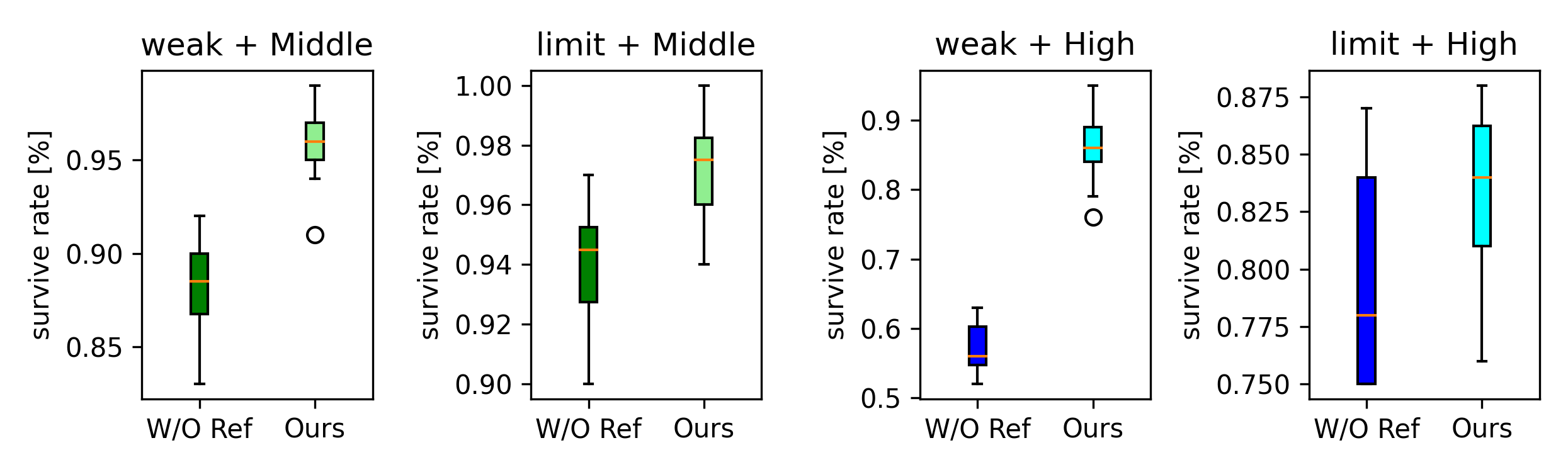}
  \caption{Survival rates of the robots in two fault status (Weak, Limit). The method trained without reflection technique is included in the comparison. Additionally, a high-speed $V_x \sim (0.8\pm 0.1)$ transition test is incorporated into this ablation experiment.}
  \label{fig:7}
\end{figure}

As shown in Fig. \ref{fig:7}. Our approach with the reflection technique demonstrates a satisfactory improvement in success rates under both electrical and mechanical failure settings. Even in the high-speed transition challenge, our policy still maintains a high success rate. This indicates that the proposed reflection technique is indeed necessary during the first training stage.

\subsection{Retention of locomotion and velocity tracking abilities}

\begin{figure}[!h]
  \centering
  \includegraphics[width=0.45\textwidth]{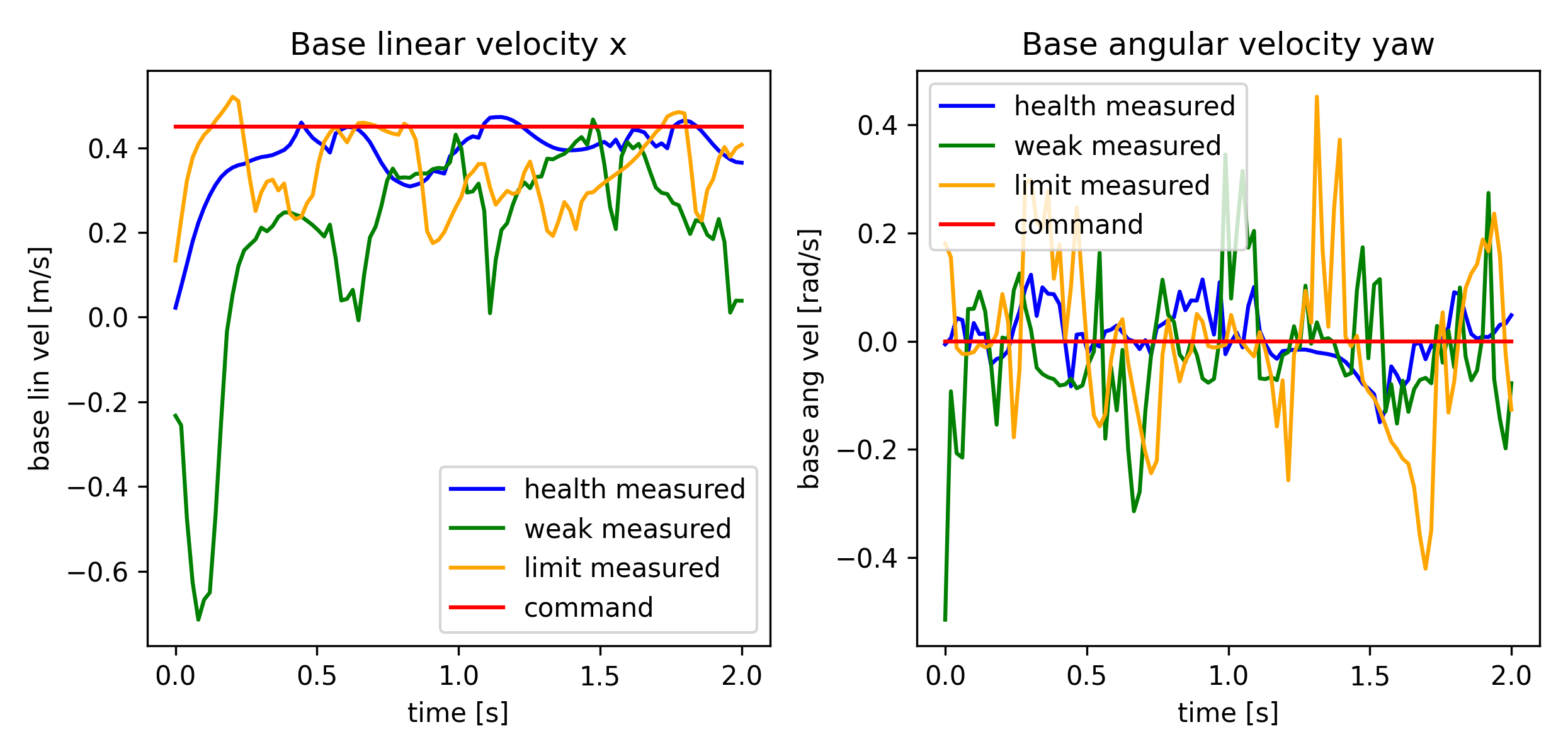}
  \caption{Linear and angular velocity measured under a 0.45m/s forward command.}
  \label{fig:8}
\end{figure}

\begin{figure}[t]
  \centering
  \includegraphics[width=0.45\textwidth]{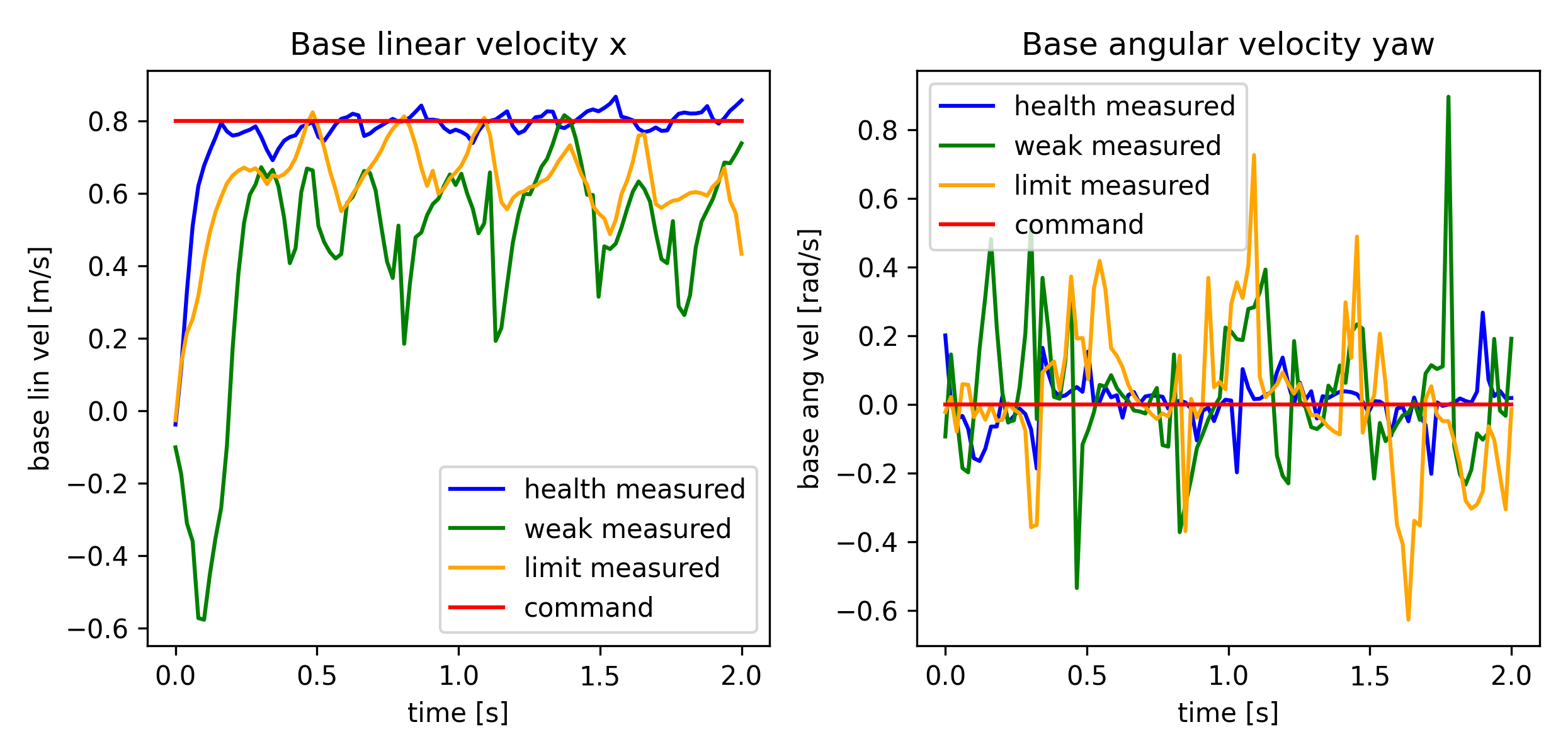}
  \caption{Linear and angular velocity measured under a 0.8m/s forward command.}
  \label{fig:9}
\end{figure}

In the third experiment, we tested the locomotion abilities under various leg conditions (health, weak, limit). All experiments involved tracking the linear velocity commands $V_x$ equals $0.45m/s$ and $0.8m/s$, Fig. \ref{fig:8} and Fig. \ref{fig:9} displays the data collected in 2 seconds. Our control scheme maintained satisfactory locomotion velocity and consistent heading across different leg conditions.

\subsection{Sim2Real Experiment}

While in simulation, both the normal and 'Mix' policies exhibit some survival rate under mechanical faults, the Sim2Real difference renders the application of the same gaits in the real world impractical. A visual examination of gait videos reveals some details.

Normal policy typically entails the locked leg making frequent contact and swiftly sliding on the ground. However, this would pose significant issues in the physical real-world. Table \ref{table2} provides statistics on the time duration of ground contact by locked leg, demonstrating our approach with minimal contact time. This is achieved through the introduction of body swing, which helps mitigate sliding and results in a more natural gait.

Finally, we deployed our strategy on a real-world solo8 robot. As shown in Fig. \ref{fig:1}, our strategy achieved zero-shot sim2real transfer, maintaining mobility in different leg status with diverse skills.

\begin{table}[h]
\caption{Feet contact floor time}
\label{table2}
\begin{center}
    \begin{tabular}{|c|c|c|c|c|}
    \hline
    Time           & $Normal$ & $Mix$ & $W/O \ Ref$ & $Ours$\\
    \hline
    High Speed               &5.7607& 6.9860& 5.5355& \textbf{4.8745} \\
    Middle Speed             &5.8428& 7.4174& 5.1119& \textbf{4.8073} \\
    \hline
    \end{tabular}
\end{center}
\end{table}

\section{CONCLUSIONS}

In this study, we investigate common leg failures encountered by quadruped robots during outdoor exploration, extending prior works by incorporating electrical failure and two joints broken scenarios. The architecture we proposed enables a variety of locomotion skills under different tasks using unified reward functions. Reflection initialization technique aids in dynamic skill transitions. Additionally, our control scheme accomplished zero-shot Sim2Real transfer on the real-world SOLO8 robot effectively managing leg failures to ensure stability and accurate planar mobility tracking of human commands. Future work involves integrating proprioceptive and visual sensors for diverse terrains, as well as implementing parallel training for a wider range of skills, thereby enhancing the overall capabilities of outdoor quadruped robots.



\bibliographystyle{IEEEtran}
\bibliography{Fmt}

\addtolength{\textheight}{-12cm}   
\end{document}